\documentclass[runningheads]{llncs}

\usepackage{eccv}

\usepackage{eccvabbrv}

\usepackage{graphicx}
\usepackage{booktabs}

\usepackage[accsupp]{axessibility}  

\usepackage[pagebackref,breaklinks,colorlinks,citecolor=eccvblue]{hyperref}

\usepackage{orcidlink}

\usepackage[table]{xcolor} 
\usepackage{multirow}
\usepackage{bbm}
\usepackage{amsfonts}
\usepackage{tikz}
\usepackage{bm}
\usepackage{wrapfig}
\usepackage{pifont}
\newcommand{\xmark}{\ding{55}}%
\usepackage{amsmath}
\DeclareMathOperator*{\argmax}{arg\,max} 

\definecolor{2dcolor}{HTML}{FFF2CC}
\definecolor{3dcolor}{HTML}{FFFFFF}
\definecolor{clipfeat}{HTML}{DAA520}
\definecolor{vasdfeat}{HTML}{1A6566}
\definecolor{ashbox}{HTML}{F5F5F5}

\newcommand{\mybox}[3]{%
{\setlength{\fboxsep}{0pt}%
\fcolorbox{black}{#3}{\phantom{\rule{#1}{#2}}}}%
}%

\DeclareRobustCommand{\ashdottedbox}{%
    \tikz[baseline=0.2ex]{ \draw[dotted, thick, draw=black, fill=ashbox] (0,0) rectangle (12pt, 8pt); }%
}

\begin{document}

\title{Open-Vocabulary and Referring Segmentation\\for 3D Gaussians Using 2D Detectors}

\titlerunning{GaussDet}

\author{Jameel Hassan, Yasiru Ranasinghe \and
Vishal Patel
}
\authorrunning{J. Hassan et al.}

\institute{Johns Hopkins University\\
\email{\{jabduls2, dransi1, vpatel36\}@jhu.edu}
}

\maketitle

\begin{abstract}
3D Gaussian Splatting (3DGS) has emerged at the forefront of 3D scene reconstruction. Extending 3DGS with language-driven, open-vocabulary understanding has gained significant attention for real-world applications such as embodied AI. Recent methods achieve this by learning an instance feature attribute and assigning semantics by distilling high-dimensional Contrastive Language-Image Pretraining (CLIP) features directly into the scene representation. However, the instance grouping mechanisms of these methods either require a predefined number of instances or suffer from noise in their bottom-up grouping strategies. Furthermore, the reliance on CLIP restricts semantic understanding to simple noun phrases, preventing complex spatial reasoning and referential expression grounding. We present \texttt{GaussDet}, a method that circumvents the need for dense CLIP features by leveraging discrete, open-vocabulary 2D object detectors with referring expression capabilities. We learn instance features for individual Gaussians to decompose the scene into 3D instance groups. By rendering these groups and aggregating semantic votes from multi-view 2D detections, we generate a robust View-Aggregated Semantic Label Distribution (VASD) for each 3D instance. This view-aggregation strategy acts as a strong regularizer, attenuating spurious labels caused by low-quality instance grouping. Our approach enables a straightforward, zero-shot extension from simple language queries to complex referential grounding. Extensive evaluations across two key tasks---open-vocabulary segmentation (LeRF-OVS, ScanNet) and referring expression grounding (Ref-LeRF)---demonstrate that \texttt{GaussDet} achieves consistent improvements over existing methods. Most notably, we achieve a substantial \textbf{16.7\% mIoU improvement} in referential grounding within a strict zero-shot setting.
  \vspace{-0.7em}
\end{abstract}

\section{Introduction}
\label{sec:intro}

Open-vocabulary capabilities have garnered significant attention across numerous computer vision domains, including image~\cite{zhao2017open, zareian2021open, li2022language} and video understanding~\cite{carion2025sam, lin2022frozen, rasheed2023fine}, as well as 3D scene understanding~\cite{peng2023openscene, nguyen2024open3dis, takmaz2023openmask3d,boudjoghra2024open, gu2024conceptgraphs}. Within the domain of neural rendering, the fast training, explicit point-based representation, and real-time rendering capabilities of 3D Gaussian Splatting (3DGS)~\cite{kerbl20233d} have driven its rapid adoption for 3D/4D reconstruction~\cite{keetha2024splatam, li2024spacetime, luiten2024dynamic}, generation\cite{yi2024gaussiandreamer, ling2024align}, and scene understanding~\cite{li2025scenesplat, zhou2024hugs}. Consequently, enabling open-vocabulary comprehension within 3DGS serves as a critical stepping stone for real-world applications such as embodied AI~\cite{shorinwa2024splat, halacheva2025gaussianvlm} and autonomous driving~\cite{hess2025splatad, zhou2024drivinggaussian}. \par

Early efforts to enhance 3DGS with language-grounding capabilities focused on incorporating a learnable language attribute to individual Gaussians \cite{qin2024langsplat, zhou2024feature, ye2024gaussian, shi2024language}. However, these strategies remained largely 2D-centric, relying on rendering language features onto 2D image planes for query evaluation. Pivoting toward a more 3D-native approach, OpenGaussian\cite{wu2024opengaussian} disentangled instance and semantic properties by first learning an instance attribute supervised via view-agnostic SAM masks~\cite{kirillov2023segment}. These instance features, combined with 3D coordinates, are clustered using $k$-means to group Gaussians into instances, with each instance group\footnote{\scriptsize The term ``instance group'' is used to refer to all Gaussians belonging to a single instance.} assigned a 2D CLIP feature via a rule-based view selection strategy. Building on this, LaGa\cite{cen2025tackling} adopts a similar instance-learning paradigm but eliminates the need to predefine the cluster count by grouping Gaussians through a 2D-3D association, clustering rendered SAM feature masks using HDBSCAN~\cite{mcinnes2017hdbscan}. Despite these differences, both methods fundamentally rely on dense CLIP features to assign semantics to 3D instances. This dependence introduces severe bottlenecks, limiting open-vocabulary querying by (1) restricting queries strictly to simple noun phrases~\cite{yuksekgonul2022and}, and (2) suffering from low semantic expressiveness due to the loss of surrounding image context when extracting CLIP features from isolated SAM masks. \par

\begin{figure}[t]
    \centering
    \includegraphics[width=\linewidth]{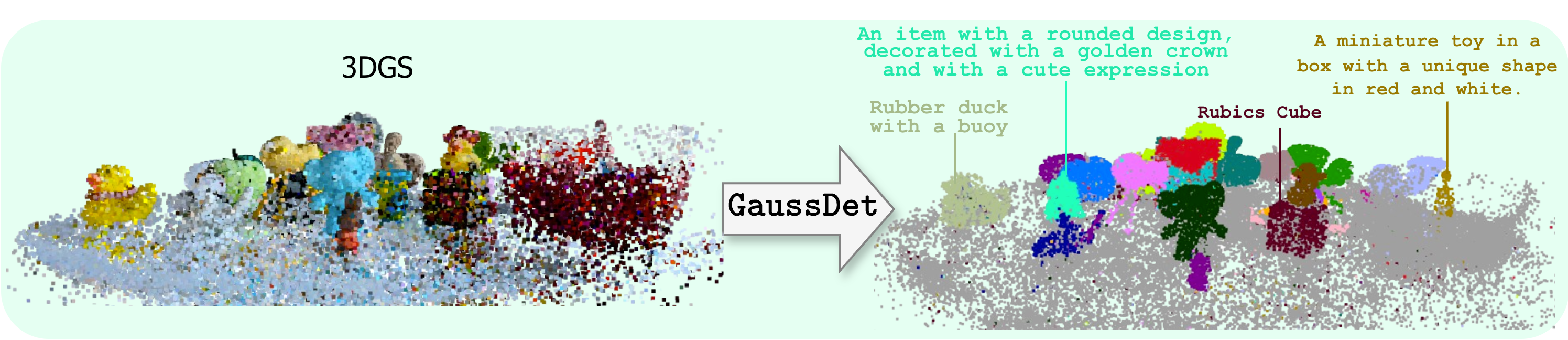}
    \caption{\textbf{Open-vocabulary and referential expression grounding with \texttt{GaussDet}.} Referential grounding examples for ``\textit{Toy cat statue}'' and ``\textit{Waldo}'' are shown.}
    \label{fig:intro}
\end{figure}

Motivated by recent breakthroughs in open-vocabulary object detection \cite{cheng2024yolo, minderer2022simple, bai2025qwen3}, we propose \texttt{GaussDet}. Our method leverages the 2D-3D associations between Gaussian instance groups and 2D bounding box detections to perform robust open-vocabulary segmentation and referential expression grounding in 3DGS. Specifically, we utilize an open-vocabulary object detector to generate discrete semantic label maps across all image frames, and follow the grouping mechanism of \cite{cen2025tackling} to obtain 3D Gaussian instance groups. We render these instance groups onto the 2D image planes and aggregate spatial votes from the discrete label maps to construct a View-Aggregated Semantic Label Distribution (VASD). 
The proposed VASD serves as a robust method to counteract the spurious semantic label votes caused by low-quality instance groupings, improving the overall semantic label quality. We present a sample visualization of our method in Fig.~\ref{fig:intro}. Our primary contributions are as follows:
\begin{itemize}
    \item We introduce a novel, 2D object detector-based approach for the semantic labeling of 3D Gaussian splats, significantly improving open-vocabulary segmentation without relying on dense CLIP features.
    \item We demonstrate the seamless extension of our pipeline to referential expression grounding in 3DGS within a zero-shot setting.
    \item Extensive evaluations show that \texttt{GaussDet} achieves substantial performance gains over existing 3DGS open-vocabulary methods, including a notable absolute increase of $16.7\%$ mIoU on the Ref-LeRF benchmark compared to the recent, per-scene optimized ReferSplat \cite{he2025refersplat}.
\end{itemize}

\section{Related Work}

\noindent\textbf{Neural Rendering:} 
The introduction of Neural Radiance Fields (NeRF) \cite{mildenhall2021nerf} marked a paradigm shift in novel view synthesis and 3D scene representation, achieving highly photorealistic renderings. While numerous subsequent works sought to improve rendering quality of NeRFs \cite{barron2021mip, barron2023zip}, the implicit continuous representation of NeRFs inherently relies on dense ray marching, leaving training and rendering computationally expensive. The recent developments in 3D Gaussian Splatting (3DGS) \cite{kerbl20233d} represents scenes explicitly using a set of anisotropic 3D Gaussians, optimized via a differentiable tile-based rasterizer. This explicit formulation achieves state-of-the-art rendering quality and real-time frame rates and provides a structured, point-like representation that is highly amenable to downstream 3D scene understanding and object-level manipulation. Consequently, 3DGS has been rapidly adopted across a wide range of real-world applications, including autonomous driving \cite{hess2025splatad, zhou2024drivinggaussian} and embodied AI \cite{shorinwa2024splat, halacheva2025gaussianvlm}.

\noindent\textbf{Open-Vocabulary Segmentation of Gaussian Splats:} 
Initial efforts to equip 3DGS with open-vocabulary understanding adopted 2D-centric strategies. Methods like LangSplat \cite{qin2024langsplat} and Feature 3DGS \cite{zhou2024feature} append high-dimensional semantic features directly to individual Gaussians for 2D query evaluation. Recent works transition to a 3D-centric approach, disentangling instance learning from semantic assignment. OpenGaussian \cite{wu2024opengaussian} groups Gaussians using $k$-means clustering on SAM-supervised instance features, but requires cluster count ($K$). A bottom-up instance aggregation is proposed in~\cite{li2025instancegaussian} to  overcome this. Meanwhile, \cite{cen2025tackling} bypasses $k$ by clustering rendered SAM mask features via HDBSCAN; however, this 2D-3D association frequently leads to under-segmentation and captures spurious masks. Crucially, these 3D-centric methods remain bottlenecked by their reliance on dense CLIP features for the final semantic assignment, which restricts queries to simple noun phrases.

\noindent\textbf{Referring Expression Segmentation:} 
Referring segmentation grounds complex, descriptive natural language queries into precise object masks. While highly successful in 2D using vision-language fusion and large multimodal models \cite{liu2024grounding, lai2024lisa} 3D referential expression grounding has traditionally focused on point clouds \cite{chen2020scanrefer, arnaud2025locate}. These methods require extensive 3D-text paired datasets to train dedicated cross-modal matching networks. In the nascent field of 3D Gaussian Splatting, ReferSplat \cite{he2025refersplat} pioneers this task by employing specialized network modules and GroundedSAM to generate pseudo-masks for per-scene training. In stark contrast to ReferSplat's reliance on per-scene optimization and pseudo-mask generation, \texttt{GaussDet} demonstrates that highly robust referential expression grounding in 3DGS can be achieved in a zero-shot manner by leveraging a 2D open-vocabulary detector within our proposed view-aggregation framework.
\vspace{-0.3em}

\section{Preliminaries}
3D Gaussian Splatting (3DGS) represents a scene as a set of anisotropic 3D Gaussians characterized by their means $\bm{\mu} \in \mathbb{R}^{3}$, a covariance matrix $\bm{\Sigma}$, color $\bm{c}$, and opacity $\bm{\sigma}$. The Gaussian parameters are optimized by rendering them to 2D image planes using a differentiable rasterization process $\bm{\psi}$, and minimizing an image-level objective. The 2D images resulting from $\bm{\psi}(\cdot)$ are computed via front-to-back alpha blending. The rendered color at pixel $\mathrm{p}$ is formulated as:
\begin{equation}
    C(\mathrm{p}) = \sum_{k=1}^{|\mathcal{O}|} \bm{c}_k \alpha_k \prod_{j=1}^{k-1}{(1 - \alpha_j),}
\end{equation}
where $\mathcal{O}$ is the depth-sorted, ordered set of Gaussians that overlap pixel $\mathrm{p}$, and $|\mathcal{O}|$ is the total number of such Gaussians. The index $k$ represents the front-to-back ranking of a Gaussian within this set. The evaluated opacity $\alpha_k$ of the $k^\text{th}$ Gaussian at pixel $\mathrm{p}$ is defined as $\alpha_k = \bm{\sigma}_k \bm{G}^{2D}_k(\mathrm{p})$, which denotes the contribution (or significance) of the Gaussian $\bm{G}_k$ to pixel $\mathrm{p}$. Here, $\bm{G}_k^{2D}(\cdot)$ represents the evaluation of the 2D-projected $k^\text{th}$ Gaussian at the specific pixel location.

Recent works based on 3DGS augment each Gaussian with an instance feature $\bm{f} \in \mathbb{R}^6$, which is then rendered into 2D feature maps $\bm{M} \in \mathbb{R}^{(H \times W \times 6)}$ via the differentiable rasterization process $\bm{\psi}(\cdot)$. These instance features $\bm{f}$ are optimized through a contrastive objective combining an intra-mask loss and an inter-mask loss, supervised by SAM instance masks.

\subsection{Disentangling Instance and Semantic Learning for Open-Vocabulary Segmentation}

Our method builds upon the paradigm of disentangled instance learning and semantic feature assignment for 3D Gaussians in open-vocabulary segmentation, as established by \cite{wu2024opengaussian,cen2025tackling}. Our approach closely follows LaGa \cite{cen2025tackling}, which learns instance features as described above and subsequently groups 3D Gaussians into coherent instances via 2D-3D associations obtained by clustering 2D instance masks. Each Gaussian instance group is assigned semantic feature descriptors by clustering the CLIP features of its associated 2D masks. Multiple semantic descriptors per instance group are maintained, motivated by the fact that the semantic representation of a 3D instance is inherently view-dependent. We refer the reader to the supplementary material for further details. \par

\textbf{Limitations:} Existing methods, including LaGa, rely heavily on the dense CLIP features of segmented 2D masks to assign semantics to 3D Gaussian instance groups. This approach introduces bottlenecks at both the instance and semantic levels. At the instance level, LaGa's grouping mechanism ---driven purely by 2D mask clustering--- is prone to the erroneous grouping of masks from distinct objects, or a mismatch between the selected masks and the 3D instance. In 3D, this yields instance groups comprising multiple objects, background elements, or floating artifacts. At the semantic level, aggregating dense CLIP features across all associated 2D masks for these low-quality 3D groups degrades the overall semantic representation. Furthermore, relying exclusively on CLIP restricts open-vocabulary querying capabilities to simple noun phrases \cite{yuksekgonul2022and}. 

In the following section, we introduce our method, \texttt{GaussDet}, designed to overcome these limitations and significantly extend the capabilities of open-vocabulary segmentation for 3D Gaussians. Focusing our contributions entirely on the semantic assignment phase, we propose a robust label generation pipeline that leverages 2D open-vocabulary object detectors to construct a View-Aggregated Semantic Label Distribution (VASD). To actively mitigate the noisy semantics resulting from low-quality instance masks and erroneous 3D groupings, we incorporate a Semantic Background Regularizer (SBR). Finally, the versatility of the employed open-vocabulary detector allows us to seamlessly extend \texttt{GaussDet} to referential expression grounding in a zero-shot manner.

\section{Method: GaussDet}
\vspace{-0.3em}
\subsection{Overall architecture}
\begin{figure}[t]
    \centering
    \includegraphics[width=0.95\linewidth]{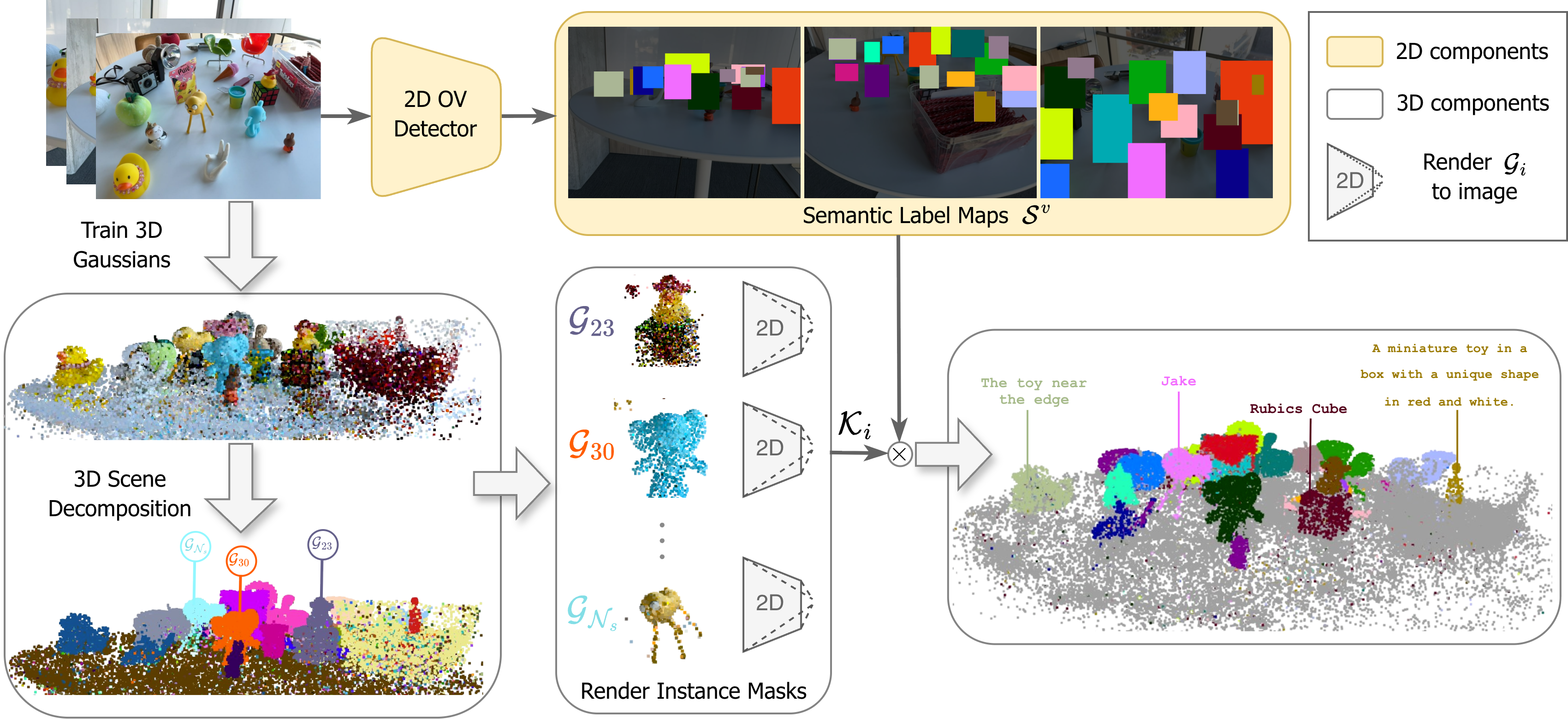}
    \caption{\textbf{\texttt{GaussDet} overall pipeline}. \raisebox{0.5ex}{\fcolorbox{black}{3dcolor}{\phantom{\rule{5pt}{2.pt}}}} We train a 3D Gaussian splat of the scene using RGB images with augmented instance features per Gaussian, which are used to decompose the scene into instance groups $\mathcal{G}_i$. Each instance group $\mathcal{G}_i$ is rendered to image frames, and the top-$K$ views $\mathcal{K}_i$ per instance group are selected. \raisebox{0.5ex}{\fcolorbox{black}{2dcolor}{\phantom{\rule{5pt}{3pt}}}} We obtain semantic label maps $\mathcal{S}^{v}$ using detections from a 2D open vocabulary object detector to generate a View-Aggregated Semantic Label Distribution (VASD) for each Gaussian via instance groups, using the semantic label maps $\mathcal{S}^{v}$ of the top-$K$ views $\mathcal{K}_i$.}
    \label{fig:pipeline}
    \vspace{-2em}
\end{figure}
Following recent approaches \cite{wu2024opengaussian, cen2025tackling}, we first augment each Gaussian with an instance feature learned via 2D instance mask supervision. We then apply the grouping mechanism from \cite{cen2025tackling} to obtain 3D instance groups $\mathcal{G} = \{ \mathcal{G}_i \in \{0, 1\}^{N} \mid i = 1, 2, \dots, \mathcal{N}_s\}$, where $N$ is the total number of Gaussians in the scene and $\mathcal{N}_s$ is the total number of instance groups. Next, we utilize an open-vocabulary object detector to generate bounding box detections $B^v$ for every frame $v$, which are subsequently used to construct a semantic label map $\mathcal{S}^v$. 

To assign a semantic label to a specific instance group, we iteratively render each group $\mathcal{G}_i$ across all views $\mathcal{V}$. During this process, we filter out Gaussians based on opacity and significance scores, and compute a rendered mask visibility score ($s_{i,v}$) for each view. These scores dictate the selection of the top-$K$ candidate views. Finally, we uplift the discrete point semantic labels from the $(x, y)$ center coordinates of the rendered Gaussians for $\mathcal{G}_i$ in these top-$K$ views to generate a View-Aggregated Semantic Label Distribution for each instance group. The overall proposed methodology of \texttt{GaussDet} is visualized in Fig.~\ref{fig:pipeline}.
\vspace{-0.7em}
\subsection{Discrete Semantic Label map generation}
Following \cite{boudjoghra2024open}, we deploy an open-vocabulary object detector to generate detections $B^v = \{(b_{i}, c_i) \mid b_i \in \mathbb{R}^4, \hspace{0.5em} c_i \in \mathbb{N}, \hspace{0.5em} i=1, \dots, D^v \}$, where $D^v$ is the number of detections in frame $v$. Each bounding box is assigned a size-based weight factor $w_i = b_{i}^H + b_{i}^W$, where $b_{i}^H$ and $b_{i}^W$ represent the height and width of the bounding box, respectively. These weights determine the rendering order of the bounding boxes when constructing the discrete semantic label map. \par

The semantic label map is represented as a 2D image $\mathcal{S}^v \in \mathbb{Z}^{(H \times W)}$, created by overlaying the bounding box labels $B^v$ at their corresponding detection locations according to the weight factor $w_i$. We initialize $\mathcal{S}^v$ with $-1$ to denote background pixels. We then sort the bounding boxes in descending order of their weights $w_i$ and populate $\mathcal{S}^v$ with the semantic labels $c_i$ across the pixel locations of each bounding box region $b_i$. The formulation of $w_i$ serves as a heuristic for occlusion modeling, operating under the assumption that smaller objects typically appear in front of larger objects along the camera's line of sight. \par

In our pipeline, we retain the semantic label for background objects as a Semantic Background Regularizer (SBR) when generating the View-Aggregated Semantic Label Distribution (VASD) for each 3D instance group. This design choice is necessitated by the lack of dense 3D instance masks obtained through the 3D scene decomposition, which we discuss in detail in Section~\ref{abl:bg-index}.
\vspace{-0.7em}
\subsection{3D instance view filtering}
\label{subsec:3d-instance-filtering}

To accurately assign semantic labels from the generated 2D maps to each instance group $\mathcal{G}_i$, we must first identify the optimal top-$K$ candidate views. Unlike 3D mask proposals derived directly from point clouds, the extracted Gaussian instance groups are often noisy, frequently capturing spurious background elements or under-segmenting instances. To address this, we first filter candidate views for semantic label uplifting using a combination of Gaussian attribute thresholding and a quantitative mask visibility score. \par

For each Gaussian instance group $\mathcal{G}_i$, we compute the set of valid rendered masks across all $\mathcal{V}$ frames. A Gaussian is considered valid in a given view only if it survives filtering based on low opacity and significance thresholds. We denote the rendered masks as:
\begin{equation}
    \mathcal{M}_i = \{ m_{i,v} \mid v \in \mathcal{V} \}.
\end{equation}

The mask visibility score $s_{i,v}$ for a given view $v$ is computed as the ratio of visible Gaussians to the total number of Gaussians in the group:
\begin{equation}
    s_{i,v} = \frac{\vert \tau_v(\mathcal{G}_i) \vert}{\vert \mathcal{G}_i \vert},
\end{equation}
where $\tau_v(\mathcal{G}_i)$ represents the subset of Gaussians in $\mathcal{G}_i$ that remain visible after filtering in view $v$, and $\vert \cdot \vert$ denotes the cardinality of the set. Finally, the top-$K$ candidate views, denoted as $\mathcal{K}_i$, are selected for each instance group by identifying the $K$ frames that maximize the mask visibility score $s_{i,v}$.

\subsection{View-Aggregated Semantic Label Distribution (VASD)}

\begin{figure}[!t]
    \centering
    \includegraphics[width=0.9\linewidth]{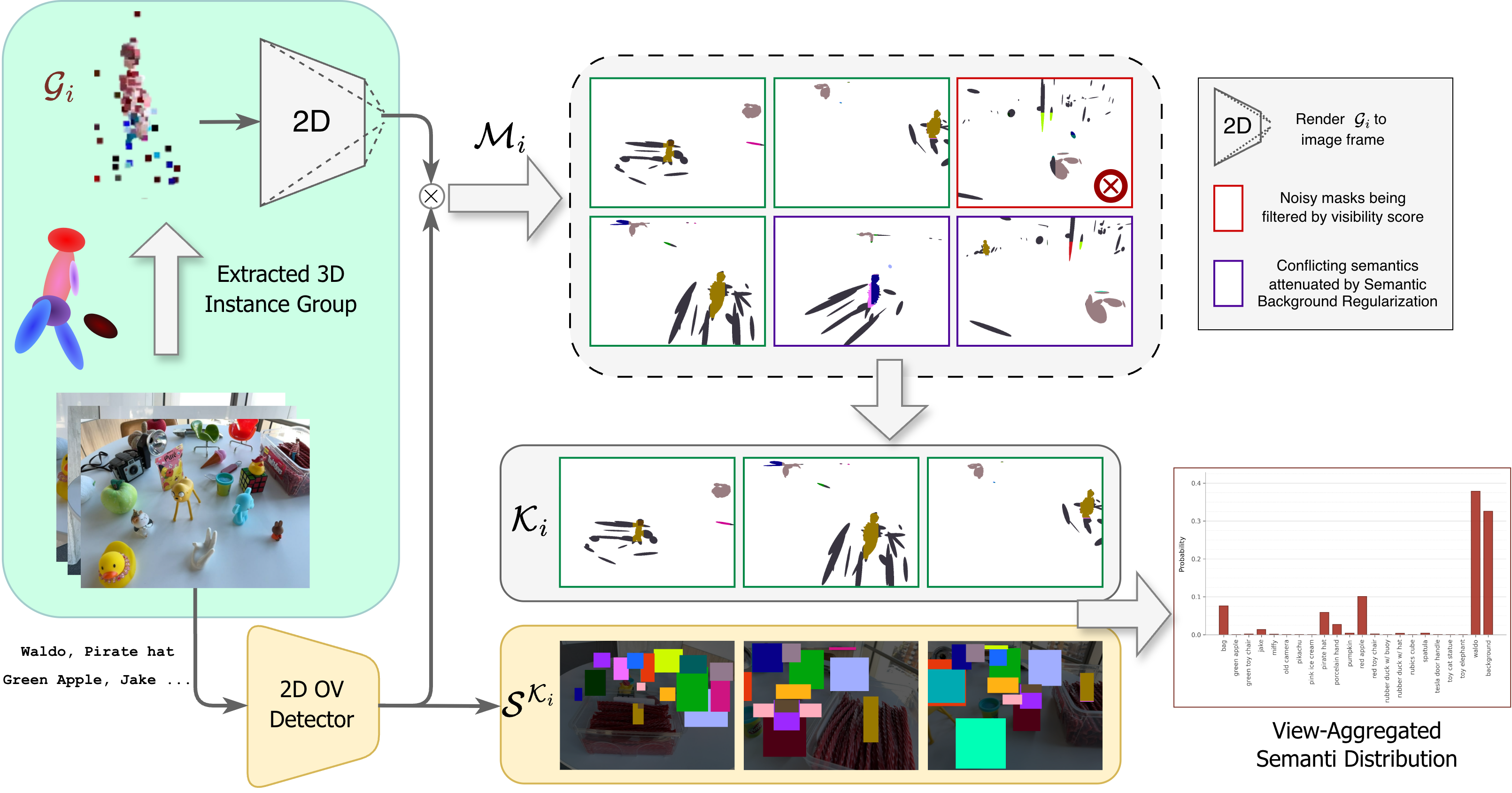}
    \caption{\textbf{View-Aggregated Semantic Label Distribution (VASD) generation.} We extract 3D instance groups $\mathcal{G}_i$ and render them to camera frames to obtained filtered masks $\mathcal{M}_i$ \ashdottedbox. The top-$K$ masks $\mathcal{K}_i$ are then chosen using the mask visibility score $s_{i,v}$. The discrete semantic label maps $\mathcal{S}^v$ of the top-$K$ views are then used to build the VASD for the instance group.}
    \label{fig:VASD}
\end{figure}

Given a 3D instance group $\mathcal{G}_{i}$, we obtain the projected 2D coordinates of the mean of each Gaussian in view $v$ by tracing it through the rasterization process, denoted as $P^{v}_{i}$. To determine the valid pixel coordinates to which each Gaussian mean projects, we remove points falling outside the image boundaries and filter out Gaussians with low opacity or significance scores in that view. We then discretize the remaining coordinates to the image grid as:
\begin{equation}
    \hat{P}^{v}_{i} = \lfloor \tau(P^{v}_{i}) \rfloor,
\end{equation}
where $\tau(\cdot)$ denotes the validity thresholding operations and $\lfloor \cdot \rfloor$ represents the discretization operation to obtain integer pixel coordinates. \par

We then aggregate the sampled semantic labels and the corresponding Gaussian significance scores across the candidate top-$K$ views. These are compiled into two sequences of arbitrary length $n$, denoted as $\ell_i$ and $q_i$:
\begin{align}
    \ell_i &= \bigoplus_{v \in \mathcal{K}_i} \mathcal{S}^v\big[\hat{P}^v_{i,x}, \hat{P}^v_{i,y}\big],\\
    q_i &= \bigoplus_{v \in \mathcal{K}_i} \mathcal{Q}^v \big[ \tau_{v}({\mathcal{G}_i}) \big],
\end{align}
where $\mathcal{K}_i$ is the set of indices for the top-$K$ views. For the semantic sequence $\ell_i$, we spatially sample the 2D discrete semantic label map $\mathcal{S}^v$. For the significance sequence $q_i$, $\mathcal{Q}^v \in \mathbb{R}^{N}$ is the 1D array of significance scores for all rendered Gaussians in view $v$, and we index it using the filtered subset of Gaussians $\tau_v(\mathcal{G}_i)$. The $\bigoplus$ operator denotes sequence concatenation across the selected views. \par

We define the View-Aggregated Semantic Label Distribution (VASD) for instance group $\mathcal{G}_i$ as the sum of the occurrences of each semantic class $c$ within $\ell_i$, weighted by the corresponding scores $q_i$ (Fig.~\ref{fig:VASD}).
\vspace{-0.5em}

\section{Experiments}
\vspace{-0.5em}
\label{sec:experiments}
We evaluate our method across two distinct task settings: open-vocabulary segmentation and referential expression grounding. For open-vocabulary segmentation, we evaluate on the LeRF and ScanNet datasets. Both datasets are evaluated using a 3D-centric approach following OpenGaussian \cite{wu2024opengaussian}: for LeRF, the relevant Gaussians for a query are selected and rendered for evaluation, whereas for ScanNet, the evaluation is performed directly at the point cloud level using ground-truth data. For referential expression grounding, we evaluate on Ref-LeRF, an extension of the LeRF dataset introduced in \cite{he2025refersplat}. Ref-LeRF annotations include attribute-specific descriptions and positional information, with an average description length spanning $13.6$ words. This benchmark strictly tests the fine-grained language grounding capabilities within 3D scenes. \par

\noindent\textbf{Implementation Details:} To generate the discrete semantic label maps for the fine-grained classes in the LeRF dataset, we utilize Qwen-VL $2.5$ (7B-Instruct) for object detection. For the ScanNet dataset, we deploy YOLO-World (X-Large). During the rendering of 3D instance groups for semantic label generation, we fix the opacity threshold to $0.1$ and the significance threshold to $0.2$ across all datasets. We employ an ensembling approach for the top-$K$ view selection across $K = \{20, 40, 60, 80, 120, \vert \mathcal{V} \vert \}$, as discussed in Section \ref{abl:topk}. All experiments are conducted on a single NVIDIA RTX 4090 GPU.

\subsection{Open-Vocabulary Segmentation in 3D}
\vspace{-0.5em}
\begin{table}[htb]
    \centering\small
    \caption{Comparison of open vocabulary segmentation mIOU(\%) in LeRF-OVS. $^\dagger$ denotes results from \cite{cen2025tackling}.}
    \vspace{-1em}
    \setlength{\tabcolsep}{6pt} 
    \resizebox{0.9\columnwidth}{!}{%
    \begin{tabular}{clccccc}
        \toprule
        & Method & Figurines & Teatime & Ramen & Kitchen & Average \\
        \midrule
        \multirow{3}{*}{\rotatebox{90}{2D}} 
        & LEGaussians & 60.3 & 44.5 & 52.6 & 41.4 & 46.9 \\
        & LangSplat & 44.7 & 65.1 & 51.2 & 44.5 & 51.4 \\
        & OccamLGS & 58.6 & 70.2 & 51.0 & 65.3 & 61.3 \\ 
        \midrule
        & OpenGaussian & 39.3 & 60.4  & 31.0 & 22.7 & 38.4 \\
        & LEGaussians$^{\dagger}$ & 31.2 & 34.5 & 17.6 & 17.3 & 25.2 \\ 
        & LangSplat$^{\dagger}$ & 25.9 & 35.6 & 29.3 & 33.5 & 31.1 \\ 
        & SAGA$^{\dagger}$ & 36.2 & 19.3 & 53.1 & 14.4 & 30.7 \\
        & OpenGaussian~\cite{wu2024opengaussian}$^{\dagger}$ & 61.1 & 59.1 & 29.2 & 31.9 & 45.3 \\
        & LaGa~\cite{cen2025tackling} & 64.1 & 70.9 & 55.6 & \textbf{65.6} & 64.1\\
        \rowcolor{blue!20}
        \multirow{-7}{*}{\rotatebox{90}{3D}} 
        & \texttt{GaussDet} & \textbf{72.5} & \textbf{76.6} & \textbf{65.7} & 64.9 & \textbf{69.9} \textcolor{teal}{(+5.8)}\\
        \bottomrule
    \end{tabular}
    }
    \vspace{-1.5em}
    \label{table:lerf-ovs}
\end{table}
Table \ref{table:lerf-ovs} compares the open-vocabulary segmentation performance of \texttt{GaussDet} on the LeRF-OVS benchmark against both 2D- and 3D-centric methods. Our method achieves an overall improvement of $5.8\%$ mIoU over LaGa, with particularly notable gains in highly cluttered scenes such as \textit{Figurines} and \textit{Ramen}. The \textit{Waldo-Kitchen} scene presents a unique challenge due to numerous heavily overlapping objects, requiring a wide variation in viewing angles for our 2D bounding box-based label maps to distinguish instances accurately. Despite this extreme viewpoints, our method's performance drop in this specific scene is marginal, demonstrating the resilience of our view-aggregation strategy.
\vspace{-0.5em}

\subsection{Open-Vocabulary Point Cloud Segmentation}
We evaluate point cloud segmentation performance against OpenGaussian and LaGa in Table \ref{table:scannet}. We report results across the standard class splits utilized in OpenGaussian, though our semantic label map generation operates only over all 19 classes simultaneously. Notably, our detector-based method demonstrates a substantial $3.3\%$ mIoU improvement on the full 19-class split compared to the \textit{officially reported} LaGa numbers. Furthermore, when compared against our reproduced baseline numbers for OpenGaussian and LaGa (which are lower than reported), \texttt{GaussDet} showcases an even more massive performance gap.
\begin{table}[!b]
    \centering\small
    \caption{Comparison of open vocabulary point cloud segmentation on ScanNet dataset. \textsuperscript{$\ddagger$} denotes numbers reproduced by us using their original code.}
    
    \setlength{\tabcolsep}{7pt} 
    \vspace{-1em}
    \resizebox{0.8\columnwidth}{!}{%
    \begin{tabular}{lcccccc} 
        \toprule
        \multirow{2}{*}{Method} & \multicolumn{2}{c}{10 classes} & \multicolumn{2}{c}{15 classes} & \multicolumn{2}{c}{19 classes} \\
        \cmidrule(lr){2-3} \cmidrule(lr){4-5} \cmidrule(lr){6-7}
         & mIoU & mAcc & mIoU & mAcc & mIoU & mAcc \\
        \midrule
        
        \color{gray} OpenGaussian & \color{gray} 38.3 & \color{gray} 55.2 & \color{gray} 30.1 & \color{gray} 48.8 & \color{gray} 24.7 & \color{gray} 41.5 \\
        \color{gray} LaGa & \color{gray} 42.6 & \color{gray} 63.2 & \color{gray} 35.5 & \color{gray} 53.5 & \color{gray} 32.5 & \color{gray} 49.1 \\
        
        OpenGaussian~\cite{wu2024opengaussian}$^{\ddagger}$ & 34.0 & 47.4 & 27.5 & 39.5 & 25.3 & 36.9 \\
        LaGa~\cite{cen2025tackling}$^{\ddagger}$ & 37.4 & \textbf{53.9} & 28.9 & 44.6 & 24.8 & 38.4 \\
        \rowcolor{blue!20}
        \texttt{GaussDet} & \textbf{38.7} & 51.3 & \textbf{35.4} & \textbf{49.9} & \textbf{35.8} & \textbf{51.0} \\
        \bottomrule
    \end{tabular}
    }
    \label{table:scannet}
\end{table} 
\vspace{-0.5em}

\subsection{Referential Expression Segmentation in 3D}
\begin{table}[htb]
    \centering\small
    \caption{Comparison of referntial grounding in 3D Gaussian Splats.} 

    \setlength{\tabcolsep}{6.5pt} 
    \vspace{-1em}
    \resizebox{0.9\columnwidth}{!}{%
    \begin{tabular}{lccccc}
        \toprule
        Method & Figurines & Teatime & Ramen & Kitchen & Average \\
        \midrule
        Grounded SAM & 16.0 & 16.9 & 14.1 & 16.2 & 15.8 \\
        LangSplat & 17.9 & 7.6 & 12.0 & 17.9 & 13.9 \\
        SPIn-NeRF & 9.7 & 11.7 & 7.3 & 10.3 & 9.8 \\
        GS-Grouping & 8.6 & 14.8 & 27.9 & 6.3 & 14.4 \\
        GOI & 16.5 & 22.9 & 27.1 & 15.7 & 20.5 \\
        ReferSplat~\cite{he2025refersplat} & 25.7 & 31.3 & 35.2 & 24.4 & 29.2 \\
        \rowcolor{blue!20}
        \texttt{GaussDet} (Zero-shot) & \textbf{58.7} & \textbf{39.2} & \textbf{52.3} & \textbf{33.3} & \textbf{45.9} \textcolor{teal}{(+16.7)}\\
        \bottomrule
    \end{tabular}
    }
    \label{table:ref-lerf}
\end{table}

By eliminating the reliance on dense CLIP features for semantic assignment, \texttt{GaussDet} inherently supports more complex textual queries. We demonstrate the benefit of this versatility by evaluating our method on the Ref-LeRF referential expression grounding benchmark in Table \ref{table:ref-lerf}. It is important to note that the baseline, ReferSplat, utilizes GroundedSAM to generate pseudo-masks for queries in the training set to optimize the referring features of Gaussians, employing dedicated network modules to learn position-aware cross-modal interactions. \texttt{GaussDet} shows a significant overall improvement of {$\bm{16.7}\%$ \textbf{mIoU}} over ReferSplat—despite operating in a strict zero-shot setting—solely by leveraging a capable open-vocabulary detector and our robust VASD assignment. Consistent with our OVS results, we observe the highest performance margins on heavily cluttered scenes such as \textit{Figurines} and \textit{Ramen}.
\begin{figure}[!t]
    \centering
    \includegraphics[width=0.85\linewidth]{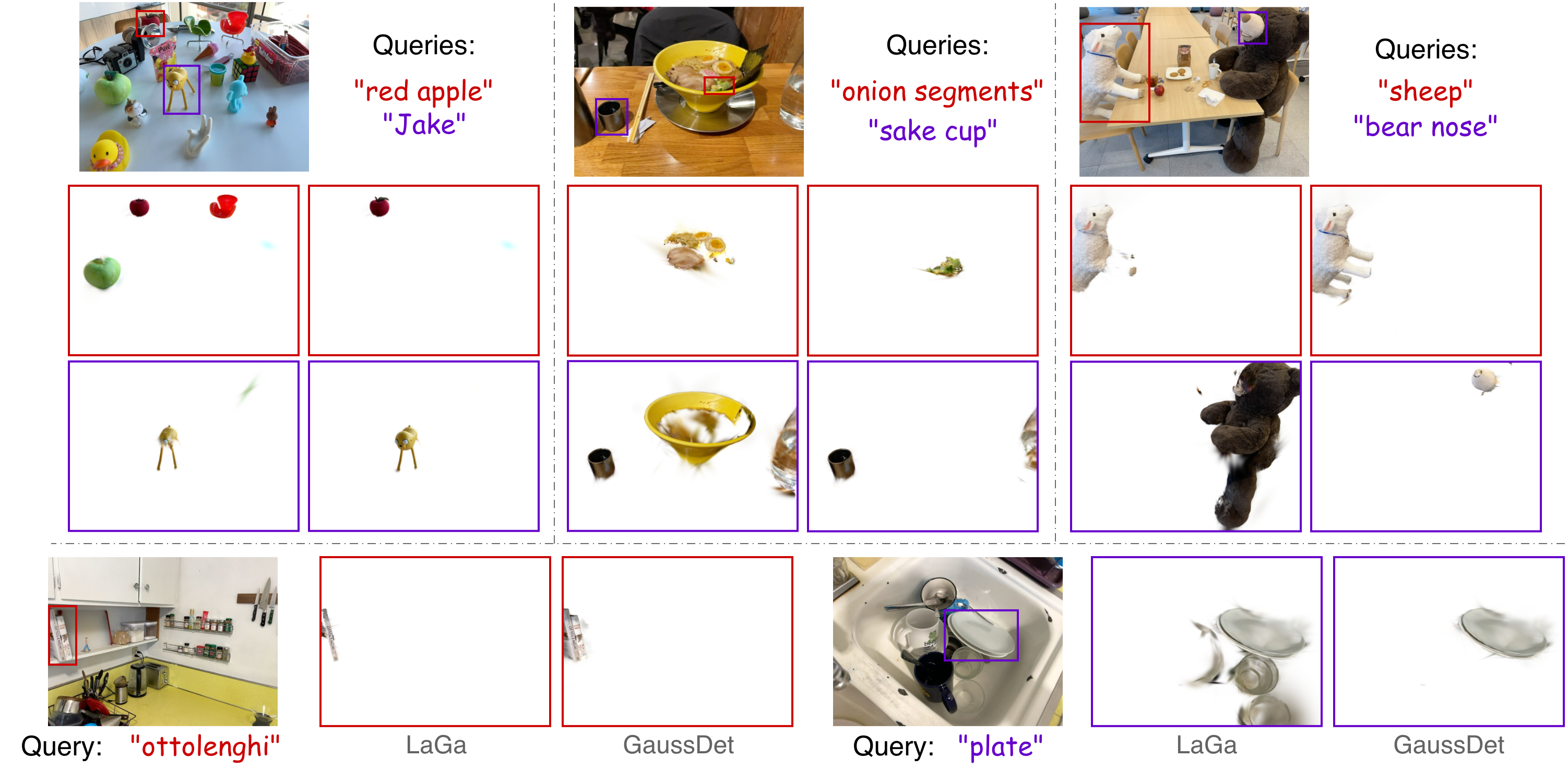}
    \caption{\textbf{Qualitative comparison on the LeRF dataset for Open-Vocabulary Segmentation.} For each query pair, the segmentation output from LaGa~\cite{cen2025tackling} (\texttt{left}) and the  output from our method (\texttt{right}) are compared. \texttt{GaussDet} yields significantly sharper object boundaries and prevents semantic bleeding into the background.}
    \vspace{-0.5em}
    \label{fig:ovs-qual}
\end{figure}

\subsection{Qualitative Results}
Visual comparisons of our open-vocabulary segmentation outputs further validate the quantitative gains over baseline methods. As illustrated in Fig.~\ref{fig:ovs-qual}, \texttt{GaussDet} yields significantly sharper segmentation boundaries compared to LaGa, evident in the clean extraction of objects like the toy ``Jake'' and the ``ottolenghi'' book. Furthermore, our method demonstrates vastly superior instance isolation. When queried with distinct items within clutter ---such as ``red apple'', ``sake cup'', ``onion segments'' and ``plate''--- LaGa frequently captures multiple adjacent instances. In contrast, our View-Aggregated Semantic Label Distribution (VASD) explicitly localizes the precise target. Our method also successfully segments queries of object parts such as  ``bear nose'', a benefit from detections on the complete image over segmented mask features.\par

\begin{figure}[!t]
\vspace{-1em}
    \centering
    \includegraphics[width=\linewidth]{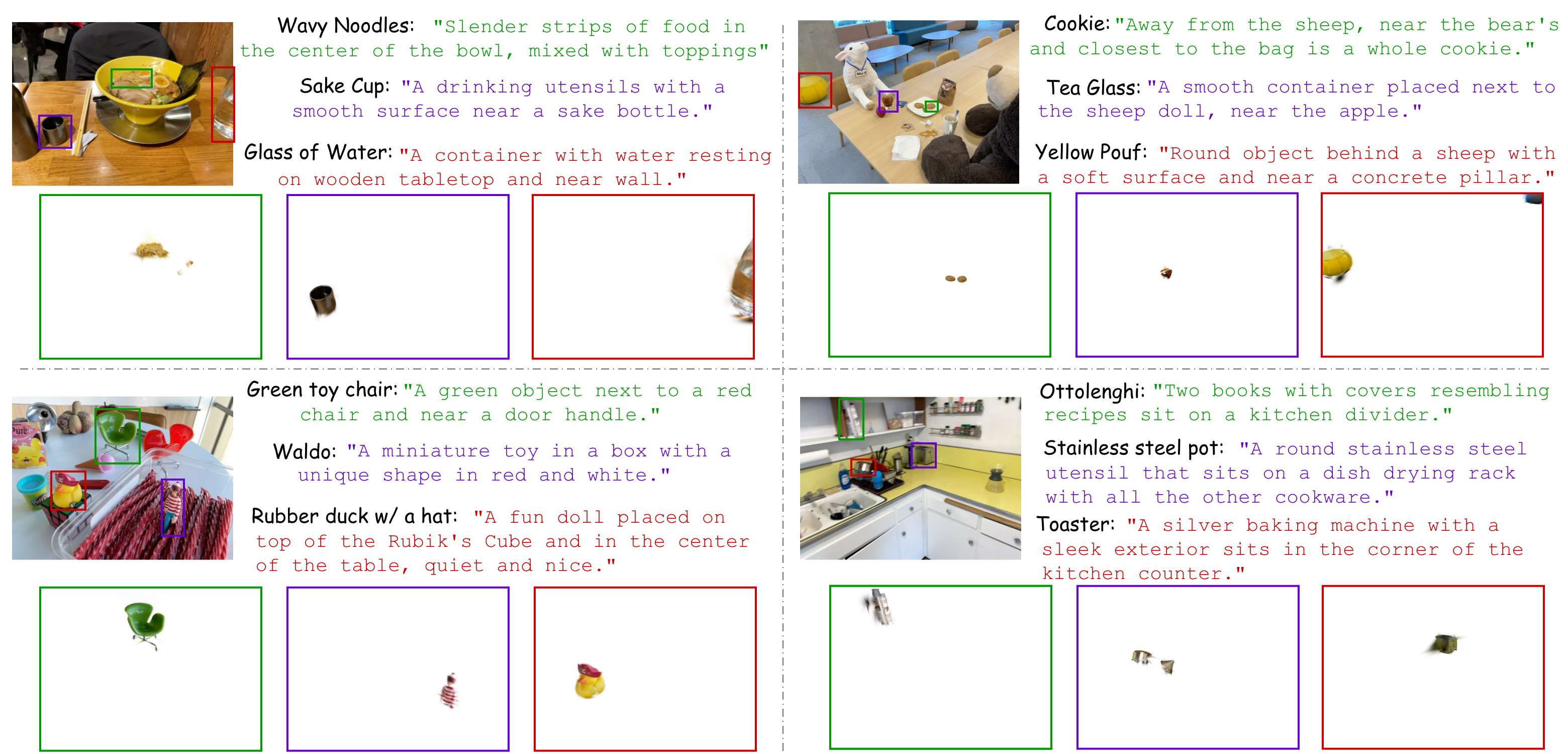}
    \caption{\textbf{Qualitative results on Ref-LeRF dataset for Referential Grounding.}}
    \label{fig:reflerf-qual}
\end{figure}

We visualize the robust zero-shot capabilities of \texttt{GaussDet} on complex referring expressions in Fig.~\ref{fig:reflerf-qual}. As seen in the qualitative examples, \texttt{GaussDet} accurately resolves queries requiring fine-grained spatial awareness (e.g., ``Rubber duck on top of the rubics cube'') and compositional attribute comprehension (e.g., ``wavy noodles''), cleanly isolating the 3D instances described by the text.

\section{Discussion and Ablations}
\subsection{Detector-based Semantic Label Assignment (VASD)}

In this section, we contrast the baseline CLIP dense feature-based semantic assignment utilized in \cite{cen2025tackling} with our proposed open-vocabulary detector-based semantic assignment (VASD). In LaGa~\cite{cen2025tackling}, 3D scene decomposition is performed by clustering the SAM mask features of the rendered instance feature attributes. Each 3D instance group is then associated to a set of 2D instance masks, and the CLIP features of these masks are extracted and clustered to assign semantics (details in the supplementary material). To establish a baseline for comparison, we compute the mean CLIP embedding of these associated SAM instance masks, calculate their cosine similarity against the text embeddings of the label space, and generate a baseline probability distribution. We then compare this CLIP-based distribution directly against the semantic label distribution generated by our proposed VASD. Further examples are provided in the supplementary. \par
\begin{figure}[!t]
    \centering
    \includegraphics[width=\linewidth]{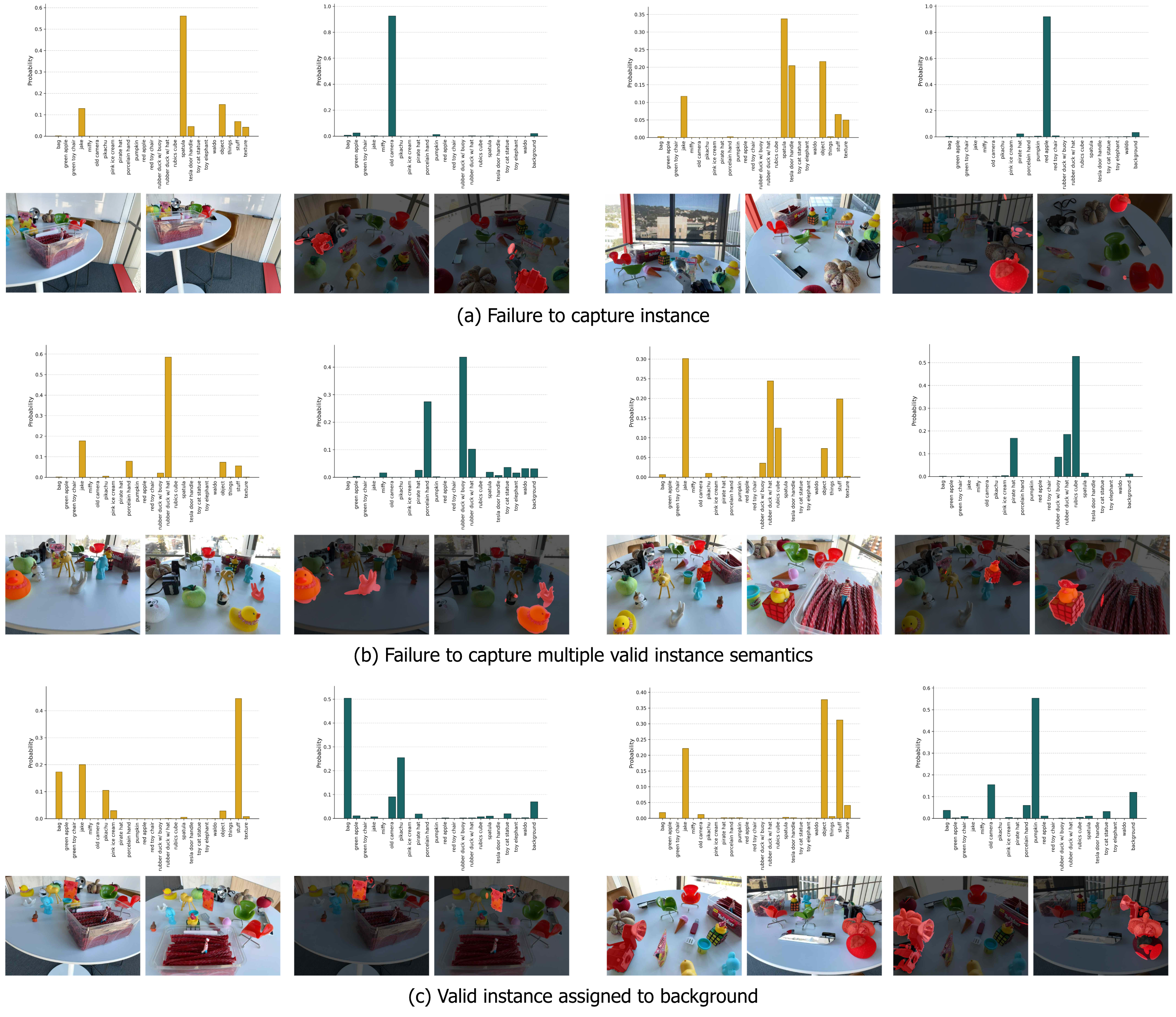}
    \caption{\textbf{Comparison of Semantic Label Distributions.} We contrast the baseline CLIP dense feature-based approach (left \mybox{3pt}{8pt}{clipfeat}) with our proposed VASD (right \mybox{3pt}{8pt}{vasdfeat}). \texttt{GaussDet} overcomes (a) spurious 2D-3D mask associations, (b) under-segmentation of distinct objects, and (c) background class domination, producing robust semantic predictions.}
    \vspace{-1em}
    \label{fig:clip-vasd}
\end{figure}
Fig.~\ref{fig:clip-vasd} highlights the advantages of our VASD detector-based approach over the CLIP-based baseline across three representative failure cases. For each subplot pair, the left plot \mybox{3pt}{8pt}{clipfeat} is the label distribution obtained using the baseline CLIP dense features alongside sample SAM instance masks (in \textcolor[HTML]{FF5C00}{orange}) associated with the 3D instance group. The right plot \mybox{3pt}{8pt}{vasdfeat} is the distribution generated by VASD, accompanied by the corresponding rendered masks overlaid on the image. \par
\textbf{Case 1: Spurious Mask Associations (Row 1, Fig.~\ref{fig:clip-vasd}a).} The baseline frequently suffers from erroneous 2D-3D associations, where the SAM instance masks assigned to a 3D instance do not actually correspond to the underlying objects (e.g., the camera and the apple). This leads to highly inaccurate semantic assignments by CLIP. Conversely, because VASD leverages discrete detector-based semantic label maps, it generates a robust and accurate semantic label distribution even if the rendered masks themselves are fragmented or incomplete. \par

\textbf{Case 2: Under-segmentation (Row 2, Fig.~\ref{fig:clip-vasd}b).} This case demonstrates scenarios where multiple distinct objects are incorrectly grouped into a single 3D instance group. Although the aggregated CLIP embeddings theoretically capture features from all constituent objects, the resulting mean distribution fails to reflect this multiplicity. For example, in the first column, the baseline assigns a negligible probability to the `\textit{porcelain hand}'. In the second column, background class probabilities incorrectly suppress the true foreground semantics. In both instances, VASD successfully generates a robust distribution that captures the multiple distinct semantics present within the group. \par

\textbf{Case 3: Background Domination (Row 3, Fig.~\ref{fig:clip-vasd}c).} We present cases where the baseline CLIP dense features align more strongly with generic or background classes (e.g., `object' or `things') rather than semantics of the foreground items. By contrast, VASD effectively isolates the foreground objects, ensuring that specific, highly descriptive semantic labels dominate the distribution.

\subsection{Semantic Label regularizer in VASD Generation}
\label{abl:bg-index}
A critical design choice in formulating the View-Aggregated Semantic Label Distribution (VASD) is the explicit inclusion of the background via the Semantic Label Regularizer (SBR). This adaptation is motivated by the inherent noisiness of 3D scene decomposition in Gaussian splatting, which frequently results in two primary contributors of noise: (1) the generation of instance groups that predominantly encompass background geometry, and (2) spurious Gaussian ``floaters'' being erroneously associated with valid instance groups. 

\begin{wrapfigure}{r}{0.44\textwidth}
    \vspace{-20pt}
    \centering
    \includegraphics[width=\linewidth]{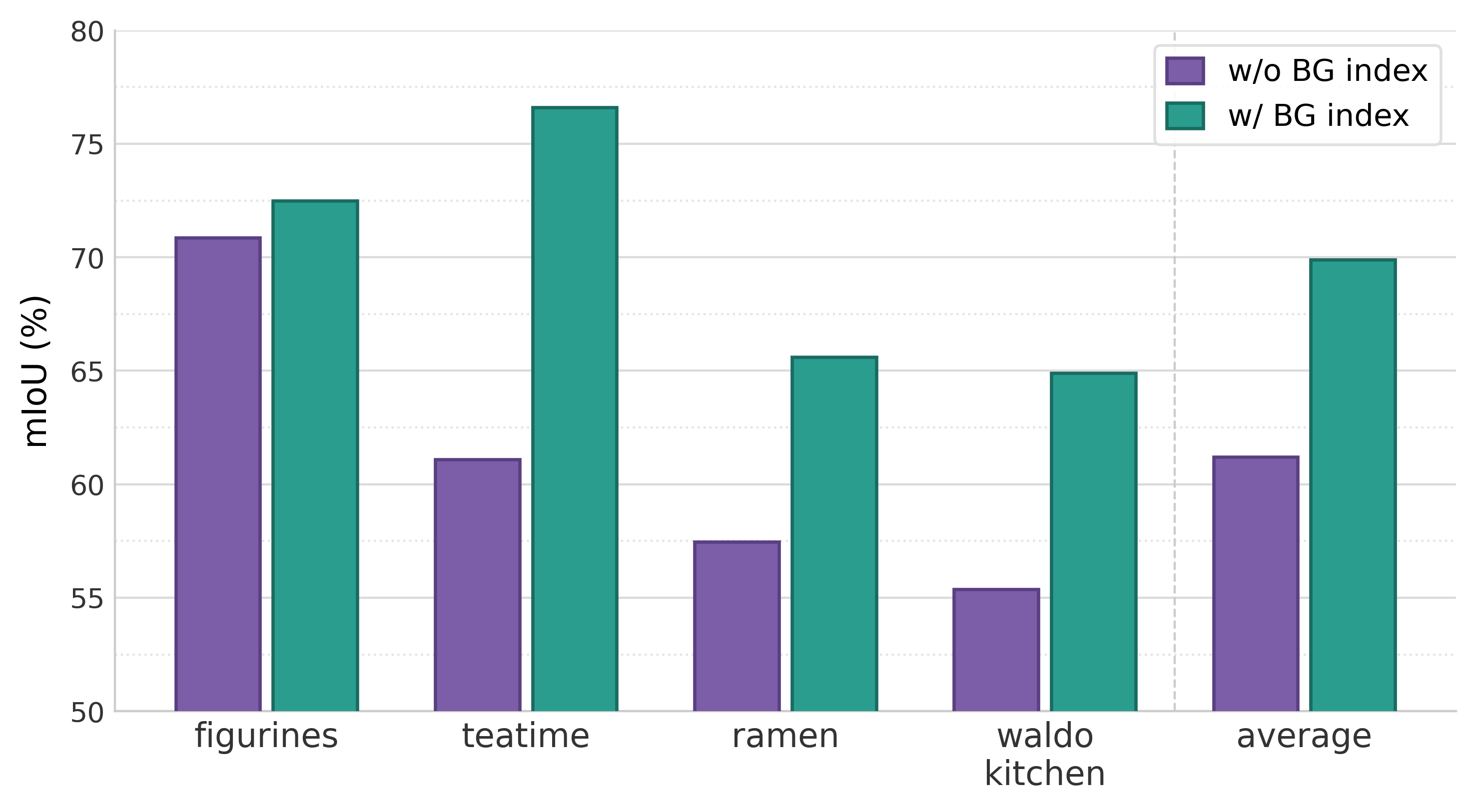}
    \caption{Impact of Semantic Label Regularizer in generating VASD.}
    \label{fig:abl-bg-index}
    \vspace{-25pt}
\end{wrapfigure}

By explicitly retaining the background index from the discrete detector label maps, SBR formulation prevents the forced, erroneous assignment of foreground semantics to groups that are actually background. This is crucial, as false positives heavily degrade downstream open-vocabulary segmentation performance. Furthermore, for spurious floaters, aggregating labels across multiple views with an active background index acts as a regularizer; the impact of an incorrect semantic assignment in one view is heavily attenuated when the floater is correctly assigned to the background in the majority of other views. As demonstrated in Fig.~\ref{fig:abl-bg-index}, ablating this feature on the LeRF dataset reveals that Semantic Label Regularization in generating the semantic label distribution yields consistent improvements across all scenes, resulting in an average mIoU increase of $6.56\%$. \par

Strategically, this formulation marks a deliberate divergence from prior open-vocabulary methods designed for point clouds, such as \cite{boudjoghra2024open}, which typically discard the background index entirely. Such methods can afford to ignore the background 
because they rely on powerful, supervised 3D mask proposal networks (e.g., trained on ScanNet) that provide strong class-agnostic geometric priors. In contrast, Gaussian splatting instances are derived through bottom-up, unsupervised 2D-3D clustering without the safety net of dense, supervised 3D object priors. Consequently, SBR explicitly becomes a necessary architectural mechanism to ensure semantic robustness in GSplat segmentation.

\subsection{Orthogonality and plug-and-play integration}
\begin{wraptable}{r}{0.6\columnwidth} 
    \vspace{-3em}
    \centering
    \caption{Transferability of VASD semantic assignment.}
    \label{tab:semantic-transfer}
    \resizebox{\linewidth}{!}{
    \begin{tabular}{lccc}
        \toprule
        Method & VASD & LeRF-OVS & ScanNet (19 classes) \\
        \midrule
        OpenGaussian & \xmark & 46.2  & 25.3 \\
        OpenGaussian + VASD & $\checkmark$ & 57.2 \textcolor{teal}{(+11.0)} & 29.3 \textcolor{teal}{(+4.0)} \\
        LaGa  & \xmark & 64.1 & 24.8 \\
        \rowcolor{blue!20}
        \texttt{GaussDet} (Ours) & $\checkmark$ & \textbf{69.9} \textcolor{teal}{(+5.8)} & \textbf{35.8} \textcolor{teal}{(+11.0)} \\
        \bottomrule
    \end{tabular}%
    }
    \vspace{-2em}
\end{wraptable}
Table~\ref{tab:semantic-transfer} demonstrates that VASD functions as a truly plug-and-play module when integrated into existing state-of-the-art open-vocabulary baselines. By delivering consistent performance gains across disparate approaches, we confirm that our semantic assignment is orthogonal to the underlying instance grouping mechanism, providing a generalized solution that effectively resolves the 3DGS semantic bottleneck.

\subsection{Effect of top-$K$ choice}
\label{abl:topk}
\begin{wrapfigure}{r}{0.5\textwidth}
\vspace{-20pt}
    \centering
    \includegraphics[width=\linewidth]{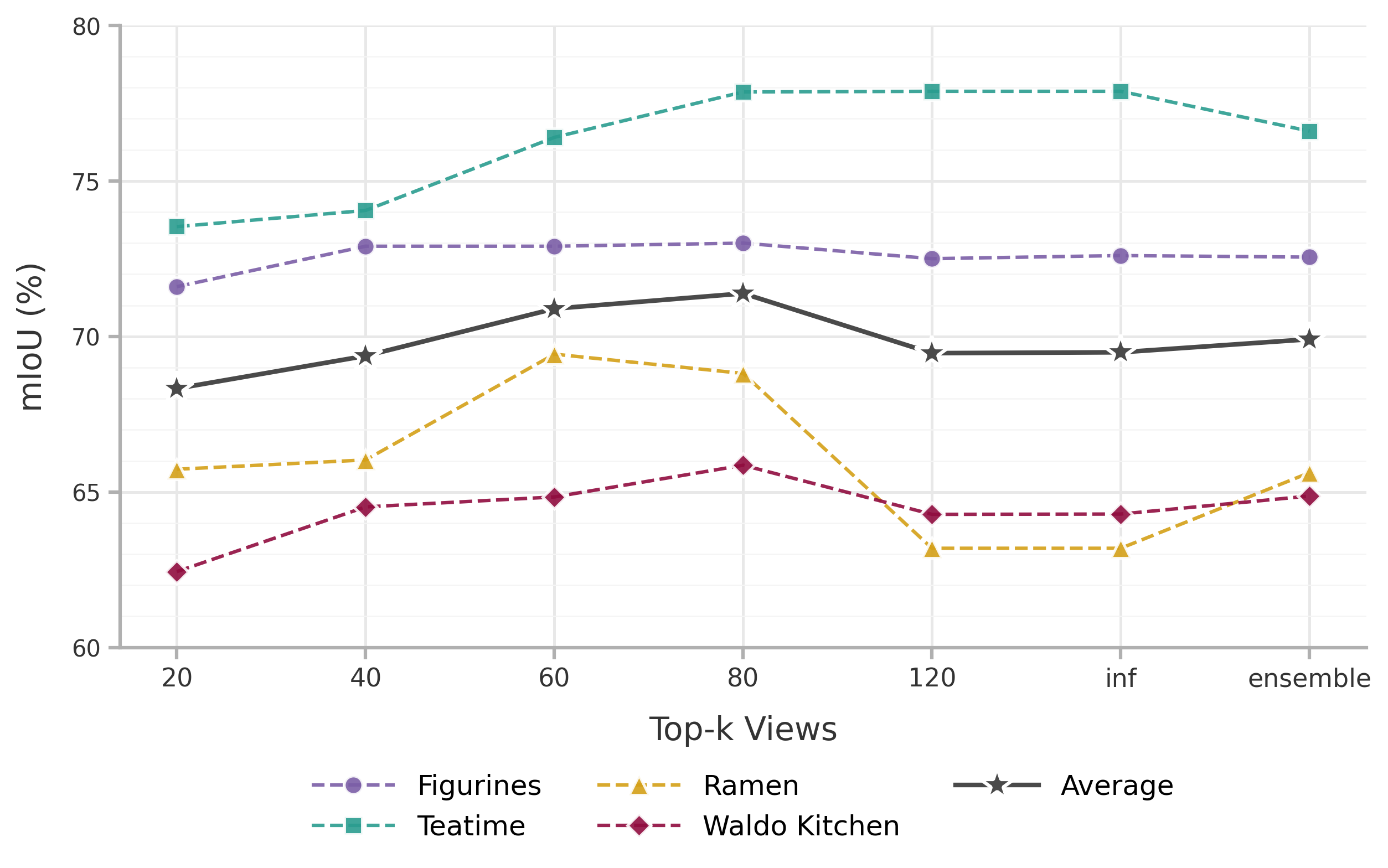}
    \caption{Influence of top-$K$ choice.}
    \label{fig:abl-topk}
\vspace{-25pt}
\end{wrapfigure}

We analyze the impact of the hyperparameter $K$, which defines the number of top-ranked views utilized to generate the view-aggregated semantic label distribution. As illustrated in Fig.~\ref{fig:abl-topk}, the optimal value for $K$ varies noticeably across different scenes. Highly cluttered environments like the \textit{Ramen} scene peak earlier at $K=60$ and degrade thereafter, likely because incorporating lower-ranked views introduces heavily occluded or noisy semantic projections. Conversely, structurally simpler scenes like \textit{Teatime} continue to benefit from broader aggregation, maintaining high performance up to $K=120$. Across all datasets, the average mIoU reaches its maximum at $K=80$ before gradually declining. \par

To address this inter-scene variance without requiring manual, scene-specific hyperparameter tuning, we introduce an ensembling approach that aggregates the semantic distributions across multiple $K$ thresholds. As shown in Fig.~\ref{fig:abl-topk}, this ensemble strategy yields robust, near-optimal performance consistently across all 
evaluated scenes. Crucially, because 3D instance groups must be rendered across all available views to compute their mask visibility scores, and the 2D discrete semantic label maps are pre-computed for all images, the marginal computational overhead of this ensembling approach is negligible. The additional cost is strictly limited to the sorting and aggregation of these pre-existing distributions.

\section{Conclusion and Limitations}
\label{sec:conclusion}

In this paper, we present \texttt{GaussDet}, a novel 3D-centric framework that advances the open-vocabulary scene understanding capabilities of 3D Gaussian Splatting (3DGS). By replacing the standard reliance on dense CLIP feature distillation with discrete semantic label maps from 2D open-vocabulary detectors, our method generates highly robust View-Aggregated Semantic Label Distributions (VASD). To mitigate the under-segmentation and noisy 2D-3D associations inherent to current 3D scene decomposition techniques, we explicitly model the background through a Semantic Background Regularizer (SBR). Consequently, \texttt{GaussDet} achieves significant improvements in both open-vocabulary segmentation and zero-shot referential expression grounding. \par

Despite these performance gains, our approach has notable limitations. First, the spatial and positional reasoning capabilities of \texttt{GaussDet} are inherently upper-bounded by the performance of the chosen 2D open-vocabulary object detector. Second, our pipeline relies on off-the-shelf 3DGS instance grouping mechanisms from which structural error could still propagate. We hope these could serve as improvements in future works.

\bibliographystyle{splncs04}
\bibliography{main}

\newpage
\appendix

\section*{ \centering \large Supplementary Material}

\section{Overview}
The following supplementary material provide additional details. A detailed description of the 3D scene decomposition method used is described in Section~\ref{supp:3d-decomp}. In Section~\ref{supp:implementation} provides further implementation details of our method, and Section~\ref{supp:qualitative} presents additional qualitative results of our method in comparison to existing works for open-vocabulary segmentation(Section~\ref{supp:ovs}), referring segmentation(Section~\ref{referring-seg}) and 3D pointcloud segmentation(Section~\ref{scannet}). We also include further ablation studies in Section~\ref{sec:further-ablations}.

\section{3D Scene Decomposition}
\label{supp:3d-decomp}
We implement our 3D scene decomposition using the method introduced in LaGa~\cite{cen2025tackling}. Inspired by prior works on 3DGS decompositions methods \cite{cen2025segment, wu2024opengaussian}, contrastive learning approach is adopted to train instance features $\mathcal{{F}} = \{ \bm{{f}}_k \in \mathbb{R}^{d}\}$, where $d$ is the dimensionality of the feature, and $k$ is the index of the Gaussian.\par

To train the instance features, a feature map $\bm{F}^v \in \mathbb{R}^{(H \times W \times d)}$ is rendered for each view $v$ using the differentiable rasterization process $\psi( \cdot )$. Each instance mask obtained using SAM~\cite{kirillov2023segment} is then assigned a instance prototype feature $\hat{\bm{f}}_{M^v}$ obtained by average pooling. The training objective is defined as:
\begin{equation}
    \mathcal{L} = \sum_{v \in \mathcal{V}} \sum_{M \in \mathcal{M}^v} \sum_{p \in \delta{(v})} {\left(1 - 2 M(p)\right) \max \left( \langle \hat{\bm{f}}_{M}, \bm{F}^v \rangle, 0 \right), }
\end{equation}
where $\delta(v)$ denotes the set of pixels in view $v$ and $M(p)$ returns true is the pixel $p$ is in mask $M$.

Let $\mathcal{M} = \bigcup_{v \in \mathcal
V} \mathcal{M}^v $ be the set of masks from all training images. These masks are then clustered based on their instance features $\hat{\bm{f}}_{M}$ using HDBSCAN~\cite{mcinnes2017hdbscan}, resulting in $\mathcal{M} = \bigcup_{i=1}^{\mathcal{N}_s} \mathcal{S}_i$, where $\mathcal{N}_s$ is the number of clusters. Each cluster $\mathcal{S}_i$ is a set of masks considered to denote a 3D instance group by establishing cross-view connections across 2D SAM masks. A 3D instance prototype feature is then computed as
\begin{equation}
    \bm{t}^{\mathcal{S}_v} = \frac{1}{\vert \mathcal{S}_i \vert} \sum_{M \in \mathcal{S}_i} \hat{f}_M.
\end{equation}
The 3D instance group $i^*$ to which the $k^\text{th}$ Gaussian is assigned is determined by identifying the 3D instance prototype with the maximum similarity to the Gaussian instance feature $\bm{f}_k$ as
\begin{equation}
    i^* = \argmax_{i} \langle \bm{f}_k, \bm{t}^{\mathcal{S}_i} \rangle.
\end{equation}
This decomposes the $N$ 3D Gaussians into instance groups $\mathcal{G} = \{ \mathcal{G}_i \in \{0, 1\}^{N} \mid i = 1, 2, \dots, \mathcal{N}_s\}$.

\section{Implementation Details}
\label{supp:implementation}
To obtain 2D instance segmentation masks, we follow previous works \cite{qin2024langsplat, wu2024opengaussian, cen2025tackling} by using a ViT-H model of SAM. The Gaussian instance feature dimension $d$ is set to 32. For LeRF-OVS dataset, we train a multi-granularity model as in LangSplat, OpenGaussian and LaGa. During inference, we obtain the average of all three levels to compute the similarity scores across the semantic label distribution. The 3DGS for each scene is trained for $30000$ iterations and the instance features are trained for a further $30000$ iterations. During inference, we set the similarity threshold to filter Gaussians as $0.5$ that removes noisy regions, then the scores are min max normalized. A 3D bilateral filtering step is applied to the resulting 3D similarity map. \par

The open-vocabulary object detector used for LeRF-OVS is Qwen-VL 2.5 Instruct, and for ScanNet, we use YOLO-World given its faster inference speed. For YOLO-World, we set the confidence threshold to be $0.08$ and apply non-maximum suppression with a threshold of $0.3$. Since Qwen-VL does not provide a confidence score, we retain all detections. To perform detections with Qwen, we pass each semantic class sequentially to the model for inference. To prevent any bias in model outputs, for each image, we randomly shuffle the semantic class order before passing them for inference. The VLM is prompted as follows
\begin{quote}
\ttfamily
prompt\_text = (\\ 
\textcolor{blue}{f}"Detect the objects corresponding to the following 
description if present: `\textcolor{orange}{\{target\_str\}}'. "\\
        \textcolor{blue}{f}"Output the results as a JSON list where each item has a key `bbox\_2d' and a value being the bounding box coordinates."\\
        )
\end{quote}
where \texttt{\textcolor{orange}{target\_str}} is the semantic class label or the referring expression.





\section{Additional Qualitative Results}
\label{supp:qualitative}

\subsection{Open-Vocabulary Segmentation}
\label{supp:ovs}
\begin{figure}[!h]
    \centering
    \includegraphics[width=0.65\linewidth]{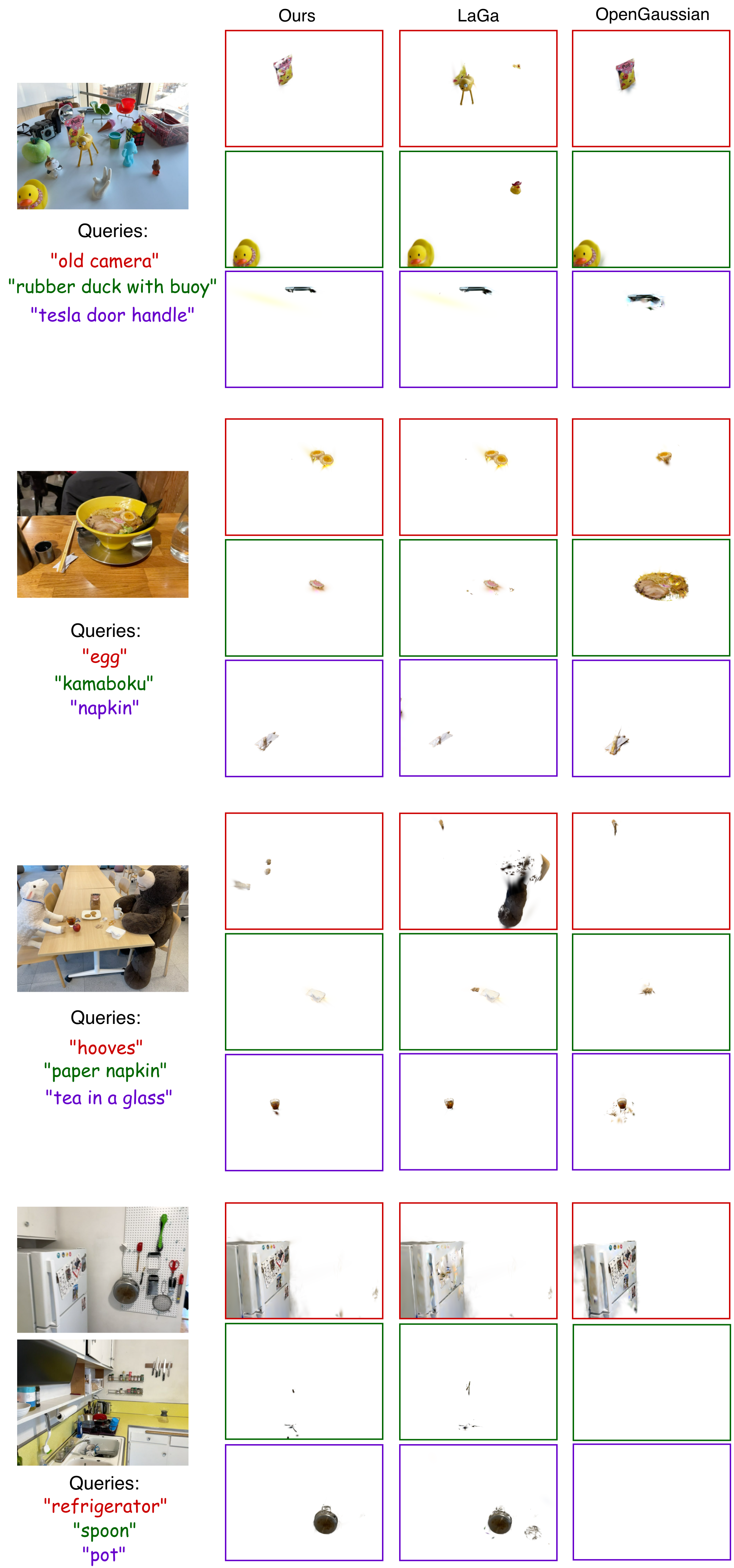}
    \caption{\textbf{Additional Open-Vocabulary Segmentation comparisons on LeRF.} For each query, we compare the segmentation outputs of OpenGaussian, LaGa, and our proposed \texttt{GaussDet}.}
    \label{fig:supp-ovs}
\end{figure}

In Figure \ref{fig:supp-ovs}, we provide further qualitative comparisons between \texttt{GaussDet} and existing works LaGa~\cite{cen2025tackling} and OpenGaussian~\cite{wu2024opengaussian} on the LeRF dataset. These examples highlight the robustness of our View-Aggregated Semantic Label Distribution (VASD) across a variety of challenging queries. We observe that LaGa and OpenGaussian tend to erroneously capture surrounding background elements rather than strictly isolating the target query. For example, LaGa captures multiple similar objects for ``\textit{rubber duck with buoy}'' query, while OpenGaussian similarly captures surrounding regions for ``\textit{apple}'' and ``\textit{kamaboku}''. Furthermore, we notice that OpenGaussian completely fails to isolate certain queries altogether. Finally, both baseline methods fail to identify fine-grained queries like ``\textit{hooves}'', whereas \texttt{GaussDet} successfully parses and segments these specific object parts.

\subsection{Referring Expression Grounding}
\label{referring-seg}
\begin{figure}[htb]
    \centering
    \includegraphics[width=\linewidth]{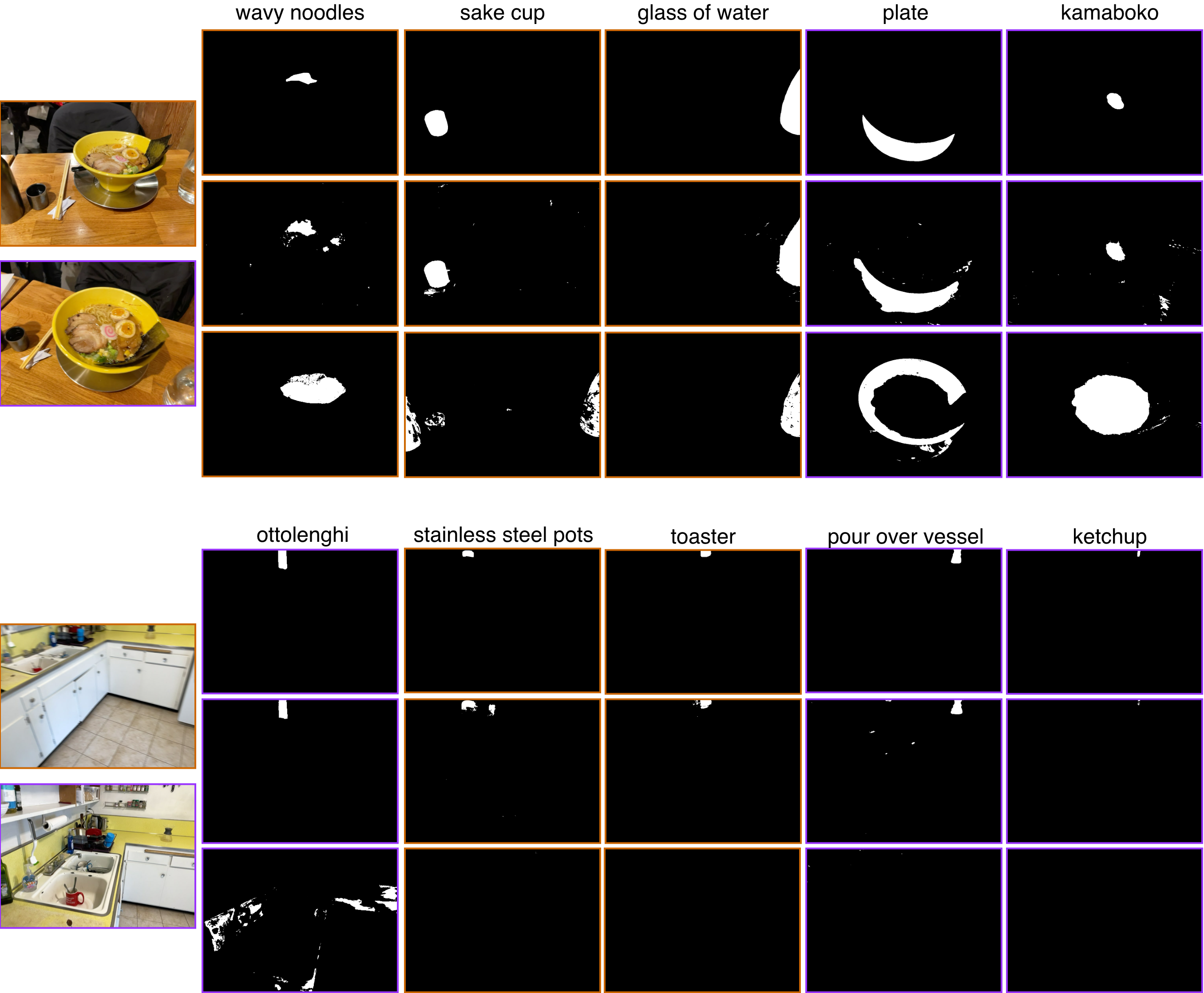}
    \caption{\textbf{Qualitative comparison on the Ref-LeRF dataset.} The first row displays the Ground Truth (GT), the second row shows the output from our method (\texttt{GaussDet}), and the third row shows the results from ReferSplat.}
    \label{fig:supp-reflerf-comp}
\end{figure}

In Figure \ref{fig:supp-reflerf-comp}, we compare the segmentation masks of \texttt{GaussDet} against the baseline ReferSplat. It is important to emphasize that our method operates in a strictly zero-shot setting, whereas ReferSplat is trained per-scene using generated pseudo-masks. Despite lacking per-scene optimization, \texttt{GaussDet} is able to isolate the target objects significantly better than ReferSplat, as demonstrated in queries such as the ``\textit{plate}'' and ``\textit{sake cup}''. Furthermore, ReferSplat sometimes completely fails to identify the queried objects altogether, as seen in the \textit{Waldo-Kitchen} example. We report visual comparisons exclusively on the \textit{Ramen} and \textit{Waldo-Kitchen} scenes, as ReferSplat only provides checkpoints and pseudo-masks for these two specific scenes.

\begin{figure}[htb]
    \centering
    \includegraphics[width=0.8\linewidth]{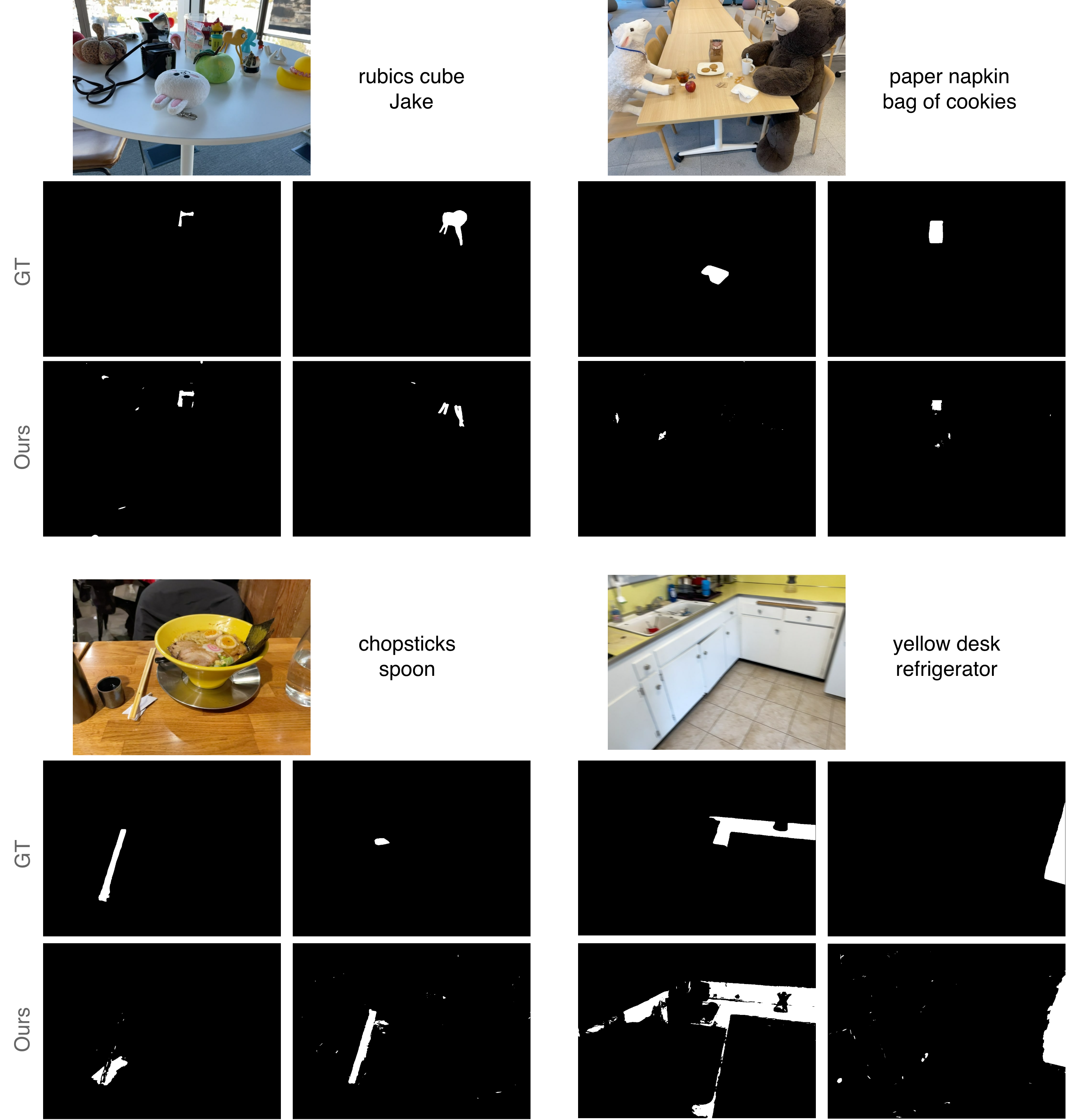}
    \caption{\textbf{Failure cases.} Our method faces challenges with heavily overlapping objects lacking multi-view coverage, leading to incomplete segmentations or residual background artifacts.}
    \label{fig:supp-reflerf-failure}
\end{figure}

\noindent\textbf{Failure Cases:} As illustrated in Figure \ref{fig:supp-reflerf-failure}, our method faces difficulties when there is not significant coverage of views for overlapping objects. In such scenarios, \texttt{GaussDet} sometimes fails to completely segment a target object, resulting in incomplete masks as seen in the ``\textit{chopsticks}'' and ``\textit{spoon}'' examples. Furthermore, the inherent limitations in the underlying 3D scene decomposition can sometimes still cause some degree of floating background artifacts to persist.

\subsection{ScanNet Qualitative Results}
\label{scannet}
\begin{figure}[!h]
    \centering
    \includegraphics[width=\linewidth]{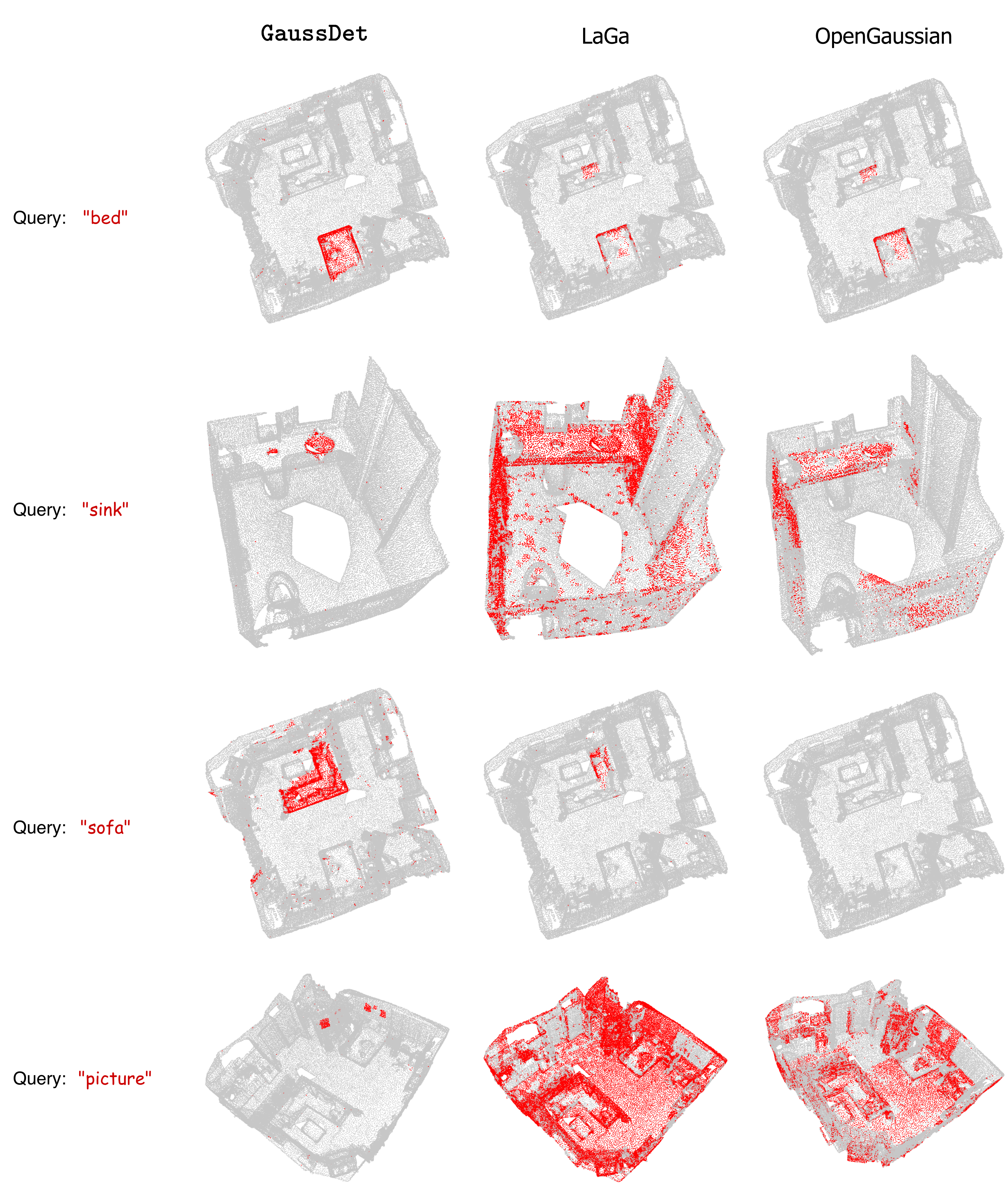}
    \caption{\textbf{Qualitative comparison on ScanNet point clouds.} We compare the segmentation outputs of OpenGaussian, LaGa, and our proposed \texttt{GaussDet}.}
    \label{fig:supp-scannet}
\end{figure}

In Figure \ref{fig:supp-scannet}, we present qualitative comparisons on the ScanNet point clouds. We observe that our segmentation masks are noticeably denser and successfully isolate background points, as demonstrated in the ``\textit{bed}'' and ``\textit{sink}'' examples. Furthermore, we notice that certain queries are completely missed by OpenGaussian, such as the ``\textit{sofa}'', which is only partially captured by LaGa. In contrast, our method obtains a highly accurate and complete segmentation of it. Finally, small objects such as the ``\textit{picture}'' are entirely missed by the baseline methods, causing severe false positive segmentations for both LaGa and OpenGaussian. Conversely, \texttt{GaussDet} performs exceptionally well at accurately localizing these small targets without introducing widespread background noise.

\section{Further Ablations}
\label{sec:further-ablations}
\subsection{Effect of VASD semantic assignment}
In order to compare the effect of the VASD semantic assignment, we compare our method with LaGa. Given the same instance grouping mechanism, we ablate the semantic expressiveness of our method on the referring segmentation benchmark in Table~\ref{tab:laga-ref} in addition to reported results in the main paper. For an exact, fair comparison of the semantic label space available to each method, we experiment our VASD semantic label distribution generation using only the query and the background label in Table~\ref{tab:single query}. We note that there is minimal difference in performance compared to using the semantic label space (Row 0) in both open-vocabulary and referring segmentation settings.
\begin{table}[htb]
    \centering
    
    \begin{minipage}{0.42\linewidth}
        \centering
        \caption{Referring segmentation using CLIP.}
        \vspace{-1em}
        \label{tab:laga-ref}
        \resizebox{0.85\linewidth}{!}{
            \begin{tabular}{lcc}
                \toprule
                Method & Ref-LeRF  \\
                \midrule
                LaGa (CLIP)  & 9.2  \\
                GaussDet (VASD)  & \textbf{45.9}\\
                \bottomrule
            \end{tabular}
        }
    \end{minipage}
    \vspace{-1.5em}\hfill
    \begin{minipage}{0.52\linewidth}
        \centering
        \caption{Using a single semantic label for VASD.}
        \vspace{-1em}
        \label{tab:single query}
        \resizebox{0.62\linewidth}{!}{
            \begin{tabular}{lccc}
                \toprule
                Method & LeRF & Ref-LeRF  \\
                \midrule
                Row 0 & 69.9 & 45.9 \\
                Row 1 & 68.9 & 46.3 \\
                \bottomrule
            \end{tabular}
        }
    \end{minipage}
    \vspace{-1.5em}
\end{table}

\subsection{Detector Uncertainty}
We study the effect of the detector with strong 2D priors and its contribution to the semantic label generation. To understand the need for a robust semantic assignment using VASD, validating that semantic understanding cannot be simply achieved by relying on the detector, we ablate the detector uncertainty. We replace the semantic label assignment using VASD as follows. We match the rendered mask’s bounding box with the bounding box that has the highest IoU overlap from the object detector outputs, and aggregate it across views for a majority vote (GaussDet + bbox). In Table~\ref{tab:detector vasd}, we observe that this significantly drops the performance, confirming that the gains are not purely from strong 2D priors. Missed/incorrect detections, and incorrect matching of mask with detections lead to degradation in performance. The drop in performance was pronounced in scenes such as “Ramen” and “Waldo Kitchen” that have highly overlapping detections. With VASD, aggregating votes across all independent Gaussians enables to generate a robust semantic distribution label, leading to significant improvement.
\begin{table}[htb]
    \centering
    \caption{Detector uncertainty analysis. VASD improvements \textcolor{teal}{teal}.}
    \label{tab:detector vasd}
    \resizebox{0.8\linewidth}{!}{
        \begin{tabular}{lccccc}
            \toprule
            Method & VASD & LeRF-OVS & Scannet19 & Ref-LeRF\\
            \midrule
            LaGa  & \xmark & 64.1 & 24.8 & 9.2\\
            GaussDet + bbox  & \xmark & 35.5 & 19.9 & 29.1\\
            \rowcolor{blue!20} GaussDet (Ours) & $\checkmark$ &\textbf{69.9} \textcolor{teal}{(+34.4)} & \textbf{35.8} \textcolor{teal}{(+15.9)} & \textbf{45.9} \textcolor{teal}{(+16.8)}\\
            \bottomrule
        \end{tabular}
    }
\end{table}

%
%

%
%
\end{document}